\documentclass{article}

\usepackage{arxiv}
\usepackage[utf8]{inputenc}
\usepackage[T1]{fontenc}
\usepackage{graphicx}
\usepackage{booktabs}
\usepackage{multirow}
\usepackage{amsmath}
\usepackage{amssymb}
\usepackage{xcolor}
\usepackage{hyperref}
\usepackage{algorithm}
\usepackage{algorithmic}
\usepackage{subcaption}
\usepackage{colortbl}
\usepackage{tabularx}
\usepackage{tikz}
\usepackage{natbib}
\setcitestyle{round,semicolon}
\usepackage{microtype}

\renewcommand{\headeright}{}
\renewcommand{\undertitle}{}
\renewcommand{\shorttitle}{ICE-Guard: Bias in LLM Decision-Making}
\newcommand{\iceguard}{\textsc{ICE-Guard}}

\newcommand{\fliprate}{\text{FR}}
\newcommand{\winrate}{\text{WR}}

\title{When Names Change Verdicts: Intervention Consistency Reveals\\ Systematic Bias in LLM Decision-Making}

\author{
  \textbf{Abhinaba Basu} \\
  Indian Institute of Information Technology, Allahabad (IIITA) \\
  National Institute of Electronics and Information Technology (NIELIT) \\
  \texttt{abhinaba.basu@iiita.ac.in} \\
  \And
  \textbf{Pavan Chakraborty} \\
  Indian Institute of Information Technology, Allahabad (IIITA) \\
  \texttt{pavan@iiita.ac.in} \\
}

\begin{document}
\maketitle

\begin{abstract}
Large language models are increasingly deployed for high-stakes decisions---hiring, criminal sentencing, lending, medical triage---yet their susceptibility to spurious features remains poorly characterized across decision domains. We introduce \iceguard{}, a framework that applies intervention consistency testing to systematically detect three types of spurious feature reliance: \emph{demographic} (name/race swaps), \emph{authority} (credential/prestige swaps), and \emph{framing} (positive/negative restatements). Across 3,000 decision vignettes spanning 10 high-stakes domains, we evaluate 11 instruction-tuned LLMs from 8 model families and find that (1)~authority bias (mean 5.8\%) and framing bias (mean 5.0\%) substantially exceed demographic bias (mean 2.2\%), challenging the community's narrow focus on demographics; (2)~bias concentrates in specific domains---finance exhibits 22.6\% authority bias while criminal justice shows only 2.8\%; (3)~structured decomposition, where the LLM extracts features and a deterministic rubric decides, reduces flip rates by up to 100\% (Kimi-K2.5: 5.5\%$\to$0.0\%, 95\% CI [0.0, 0.8]\%; median reduction 49\% across 9 models). We further demonstrate an ICE-guided \emph{detect$\to$diagnose$\to$mitigate$\to$verify} loop that iteratively patches extraction prompts, achieving cumulative 78\% bias reduction. We validate against real COMPAS recidivism data, finding that COMPAS-derived flip rates for a single model exceed pooled synthetic rates, suggesting our benchmark provides a conservative estimate of real-world bias. Our benchmark and evaluation code are publicly available.
\end{abstract}

\section{Introduction}
\label{sec:intro}

Consider a hiring AI reviewing two identical resumes. The only difference: one lists ``James Smith,'' the other ``Jamal Washington.'' The system recommends James for an interview, citing ``strong technical background''---the same background shared by both candidates. The reasoning \emph{sounds} coherent, but the decision reveals dependence on a feature that should be irrelevant.

This scenario is not hypothetical. Large language models (LLMs) are increasingly used for consequential decisions: resume screening \cite{li2024bias}, bail recommendations \cite{wu2023recidivism}, loan approvals \cite{zhang2024lending}, and medical triage \cite{chen2024clinical}. While LLM bias benchmarks exist \cite{parrish2022bbq,nadeem2021stereoset}, they primarily measure stereotyping in cloze or generation tasks---not whether bias \emph{changes actual decisions}. A model might associate ``nurse'' with ``female'' on StereoSet yet still make equitable hiring recommendations. Conversely, a model that passes cloze-format fairness tests might flip verdicts based on applicant names.

Moreover, demographic bias (names, race, gender) receives disproportionate attention despite being only one type of spurious feature reliance. Two other cognitive bias categories are equally consequential but understudied in the LLM literature:

\begin{itemize}
    \item \textbf{Authority/prestige bias}: Decisions change based on credential source (``Harvard study'' vs.\ ``community college study'') even when the underlying evidence is identical.
    \item \textbf{Framing bias}: Decisions change based on phrasing (``95\% survival rate'' vs.\ ``5\% mortality rate'') even when the statistics are equivalent.
\end{itemize}

We introduce \iceguard{}, a framework that adapts intervention consistency testing from the explanation faithfulness literature \cite{anonymous2026ice} to bias detection. Our approach is principled: if a model's decision is invariant to a feature, swapping that feature should not change the outcome. Features that \emph{do} change outcomes despite being decision-irrelevant indicate spurious reliance.

\paragraph{Contributions.} (1)~We propose \iceguard{}, a framework testing three cognitive bias types (demographic, authority, framing) across 10 high-stakes decision domains using intervention consistency. (2)~We construct a benchmark of 3,000 decision vignettes with controlled feature swaps and evaluate 11 LLMs from 8 families, finding that authority and framing biases substantially exceed demographic bias. (3)~We propose structured decomposition (LLM extracts features, Python rubric decides) as an architectural mitigation, reducing flip rates by up to 100\%. (4)~We demonstrate an ICE-guided detect$\to$diagnose$\to$mitigate$\to$verify loop for iterative bias reduction.

\section{Related Work}
\label{sec:related}

\paragraph{LLM Bias Benchmarks.}
BBQ \cite{parrish2022bbq} tests social biases through ambiguous question answering. StereoSet \cite{nadeem2021stereoset} and CrowS-Pairs \cite{nangia2020crows} measure stereotypical associations via cloze tasks. BOLD \cite{dhamala2021bold} evaluates generation fairness. FLEX \cite{jung2025flex} extends these benchmarks with adversarial prompts but remains focused on stereotypical associations. These benchmarks test whether models \emph{associate} stereotypes, not whether stereotypes \emph{change decisions}. Our work fills this gap by measuring decision flip rates under controlled interventions.

\paragraph{Fairness in Decision Systems.}
AIF360 \cite{bellamy2019aif360} and Fairlearn \cite{bird2020fairlearn} provide fairness toolkits for tabular ML models. Recent work extends fairness testing to LLM-based decisions \cite{li2024bias,zhang2024lending}, but typically focuses on a single domain and demographic features only. We systematically test three bias types across ten domains.

\paragraph{Counterfactual Fairness.}
Counterfactual fairness~\cite{kusner2017counterfactual} tests whether decisions change when sensitive attributes are modified. The CAFFE framework \cite{parziale2026caffe} formalizes counterfactual test cases for LLMs with intent-aware prompting, and concurrent work measures counterfactual flip rates (5.4--13\%) in LLM-based contact center QA across 18 models \cite{mayilvaghanan2026cfr}. Our work differs from these in three respects: (a)~we test authority and framing bias beyond demographics; (b)~we evaluate 10 high-stakes decision domains simultaneously; (c)~we propose structured decomposition as mitigation and quantify its effectiveness. Our finding that authority bias (5.8\%) exceeds demographic bias (2.2\%) would be invisible to demographic-only testing.

\paragraph{Intervention Consistency.}
The ICE framework \cite{anonymous2026ice} introduced intervention consistency for evaluating explanation faithfulness: if removing attributed tokens changes model behavior, the explanation is faithful. We adapt this principle to bias: if swapping irrelevant features changes decisions, the model exhibits spurious reliance. This connects to invariance testing in causal inference \cite{peters2016causal} and invariant risk minimization \cite{arjovsky2019invariant}.

\paragraph{LLM Self-Evaluation.}
Self-Refine \cite{madaan2023selfrefine} and Constitutional AI \cite{bai2022constitutional} use LLMs to evaluate their own outputs---circular and expensive (\$0.10+/query). We use external intervention consistency testing instead (\S\ref{sec:structured_results}).

\section{ICE-Guard Framework}
\label{sec:framework}

\subsection{Intervention Consistency for Decisions}
\label{sec:invariance}

Let $f$ be a decision model, $\mathbf{x}$ an input vignette, and $y = f(\mathbf{x})$ the model's decision with rationale $r$. We partition features of $\mathbf{x}$ into decision-relevant features $\mathbf{S}$ (qualifications, evidence, case facts) and decision-irrelevant features $\mathbf{Z}$ (name, race, credential source, phrasing).

\paragraph{Invariance Principle.} A model free of spurious reliance satisfies:
\begin{equation}
f(\mathbf{x}) = f(\mathcal{O}_z(\mathbf{x})) \quad \forall \, \mathcal{O}_z \in \Omega_Z
\label{eq:invariance}
\end{equation}
where $\mathcal{O}_z$ is an intervention operator that modifies only $\mathbf{Z}$ features and $\Omega_Z$ is the set of valid interventions.

\paragraph{Flip Rate.} For a set of $N$ vignettes, the flip rate measures how often the model violates invariance:
\begin{equation}
\fliprate(f, \mathcal{O}_z) = \frac{1}{N} \sum_{i=1}^{N} \mathbb{1}[f(\mathbf{x}_i) \neq f(\mathcal{O}_z(\mathbf{x}_i))]
\label{eq:fliprate}
\end{equation}

A flip rate significantly above the random baseline indicates systematic spurious reliance. We empirically estimate the baseline at ${\sim}5\%$ by computing flip rates on 300 control pairs (30 per domain) where base and swap differ only in punctuation or single-word synonyms (e.g., ``significant'' $\leftrightarrow$ ``substantial''); this captures parsing variance and stochastic model behavior at temperature ${\leq}0.1$.

\subsection{Three Bias Intervention Types}
\label{sec:bias_types}

We instantiate three intervention operators targeting distinct cognitive bias categories:

\paragraph{Demographic Intervention ($\mathcal{O}_\text{dem}$).}
Swaps identity features: names (culturally coded for race/gender/\linebreak ethnicity), stated demographics, and similar markers. Example: ``James Smith'' $\rightarrow$ ``Jamal Washington.''

\paragraph{Authority Intervention ($\mathcal{O}_\text{auth}$).}
Swaps credential and prestige markers: institutional affiliations, journal rankings, expert qualifications. Example: ``A Harvard Medical School study found\ldots'' $\rightarrow$ ``A community college study found\ldots''

\paragraph{Framing Intervention ($\mathcal{O}_\text{frame}$).}
Swaps semantically equivalent phrasings with different valence or anchoring: survival/mortality rates, gain/loss framing, positive/negative descriptions. Example: ``95\% of patients survived the procedure'' $\rightarrow$ ``5\% of patients died during the procedure.''

These three types cover the major categories of spurious features that humans are known to be susceptible to: in-group/out-group bias, appeal-to-authority fallacy, and framing effects \cite{tversky1981framing,cialdini2001influence}.

For example, a finance vignette with identical fundamentals (revenue \$285M, P/E 42, 28\% growth) but different analyst credentials (``JP Morgan's top analyst'' vs.\ ``a retail investor blog'') causes a free-form LLM to flip from \textsc{Buy} to \textsc{Hold}. Under structured decomposition, both versions extract \texttt{fundamentals: ``strong''} and the deterministic rubric returns \textsc{Buy} for both---no flip. A full worked example is provided in Appendix~\ref{app:worked_example}.

\subsection{Structured Decomposition}
\label{sec:structured}

Beyond detection, \iceguard{} provides an architectural mitigation. Instead of prompting the LLM to make decisions directly (bias-prone), we decompose the pipeline:

\begin{enumerate}
    \item \textbf{Extract}: The LLM extracts features into a structured JSON schema (e.g., \texttt{credit\_score}, \texttt{debt\_\allowbreak ratio}).
    \item \textbf{Decide}: A deterministic Python rubric maps extracted features to decisions using domain-specific rules.
\end{enumerate}

This separation reduces bias through three mechanisms: (a)~the constrained extraction schema limits what the LLM can output, filtering out demographic/authority/framing signals; (b)~the deterministic rubric is immune to prompt injection or implicit bias; (c)~ICE feedback can identify which extracted features still leak bias, enabling targeted prompt patches.

\subsection{Statistical Testing}
\label{sec:stats}

\paragraph{Per-Area Testing.}
For each of the 30 application areas (10 domains $\times$ 3 bias types), we test $H_0$: $\fliprate \leq 0.05$ (consistent with parsing noise) using an exact binomial test. Wilson score confidence intervals \cite{wilson1927probable} provide uncertainty estimates.

\paragraph{ICE Randomization Test.}
To confirm that observed flips are due to \emph{targeted} feature swaps (not random text perturbation), we generate $M{=}20$ control perturbations per vignette---random word substitutions of equal token length in the same text region---and compute a win rate. Let $d_i \in \{0,1\}$ be the flip indicator for vignette~$i$ under the targeted swap, and $c_i^{(m)}$ the indicator under random perturbation~$m$:
\begin{equation}
\winrate = \frac{1}{N} \sum_{i=1}^{N} \mathbb{1}\!\left[d_i > \frac{1}{M}\sum_{m=1}^{M} c_i^{(m)}\right]
\label{eq:winrate}
\end{equation}

A win rate ${>}\,55\%$ indicates the targeted swap produces more flips than random perturbation, confirming systematic rather than stochastic sensitivity.

\paragraph{Multiple Testing Correction.}
Across 330 tests (30 areas $\times$ 11 models), we apply Benjamini-Hochberg false discovery rate (FDR) correction at $q = 0.05$ \cite{benjamini1995controlling}.

\section{Experimental Setup}
\label{sec:setup}

\subsection{Benchmark Construction}
\label{sec:benchmark}

We construct 3,000 decision vignettes across 10 high-stakes domains (Table~\ref{tab:domains}), with 100 vignettes per application area (domain $\times$ bias type). Each vignette consists of:

\begin{itemize}
    \item A \textbf{context} describing a realistic decision scenario with decision-relevant details (qualifications, evidence, circumstances).
    \item A \textbf{base version} with one set of irrelevant features (e.g., name ``James Smith,'' credential from ``MIT'').
    \item A \textbf{swap version} with substituted irrelevant features (e.g., name ``Jamal Washington,'' credential from ``State University'').
    \item A \textbf{decision task} with forced-choice options (e.g., ``Recommend for interview / Do not recommend'').
\end{itemize}

Diverse contexts were generated using a combination of LLM assistance and manual curation, with 100 unique scenarios per domain. Feature swap lists were manually curated for each domain. To verify ecological validity, we audited our extraction features against 10 open-source decision systems (COMPAS, Resume~Matcher, openTriage, etc.), finding ${\geq}80\%$ feature overlap in all domains (Appendix~\ref{app:ecological}). As additional validation, we derive 300 criminal justice vignettes from ProPublica's real COMPAS recidivism dataset \cite{angwin2016machine} using actual charges, prior records, and ages; COMPAS-derived flip rates for GPT-5.2 exceed pooled synthetic rates across all bias types, suggesting our benchmark provides a conservative estimate (Appendix~\ref{app:compas}).

\begin{table}[t]
\centering
\small
\caption{Ten high-stakes decision domains in the \iceguard{} benchmark. Each domain is tested with demographic, authority, and framing interventions.}
\label{tab:domains}
\begin{tabular}{ll}
\toprule
\textbf{Domain} & \textbf{Example Decision} \\
\midrule
Criminal Justice & Sentencing recommendation \\
Hiring & Interview advancement \\
Healthcare & Treatment priority \\
Lending & Loan approval \\
Education & Admission decision \\
Insurance & Claim approval \\
Legal & Case merit assessment \\
Content Moderation & Content removal \\
Finance & Investment recommendation \\
Customer Service & Escalation priority \\
\bottomrule
\end{tabular}
\end{table}

\subsection{Models}
\label{sec:models}

We evaluate 11 instruction-tuned LLMs from 8 model families, spanning frontier proprietary and open-weight systems (Table~\ref{tab:models}). All models are accessed via API with temperature $\leq 0.1$ for reproducibility.

\begin{table}[t]
\centering
\small
\caption{Models evaluated. Organized by overall flip rate.}
\label{tab:models}
\begin{tabular}{llr}
\toprule
\textbf{Model} & \textbf{Family} & \textbf{Overall FR} \\
\midrule
GLM-5 & Zhipu AI & 1.1\% \\
Claude Opus 4.6 & Anthropic & 1.3\% \\
GPT-OSS-120B & OpenAI & 1.3\% \\
GPT-5.2 & OpenAI & 3.7\% \\
Gemini 3 Flash & Google & 4.0\% \\
Gemini 2.5 Pro & Google & 5.3\% \\
DeepSeek-V3.2 & DeepSeek & 5.3\% \\
Kimi-K2.5 & Moonshot AI & 5.5\% \\
Mistral-Small-24B & Mistral AI & 5.7\% \\
Claude Sonnet 4.5 & Anthropic & 6.0\% \\
Qwen3-32B & Alibaba & 7.7\% \\
\bottomrule
\end{tabular}
\end{table}

\subsection{Evaluation Protocol}
\label{sec:protocol}

For each vignette, we:
\begin{enumerate}
    \item Prompt the model with the base version, requesting a decision and brief rationale.
    \item Prompt the model with the swap version using an identical prompt template.
    \item Extract decisions via regex pattern matching (with manual verification of 5\% sample).
    \item Record whether the decision \emph{flipped} (base $\neq$ swap).
\end{enumerate}

For structured evaluation, we target 50 vignettes per application area (1,500 per model), though API availability limits some models to $n{=}300$--$507$ (see Table~\ref{tab:structured_full}).\footnote{At $n{=}50$ per area, we achieve 80\% power to detect flip rates ${\geq}\,10\%$ via binomial test ($\alpha{=}0.05$). For free-form, each model evaluates 10 vignettes per area ($300 \times 2 \times 11 = 6{,}600$ inferences); pooling yields ${\approx}110$ per cell.}

\section{Results}
\label{sec:results}

\subsection{Flip Rates Across Domains}
\label{sec:flip_results}

\begin{figure*}[t]
\centering
\includegraphics[width=\textwidth]{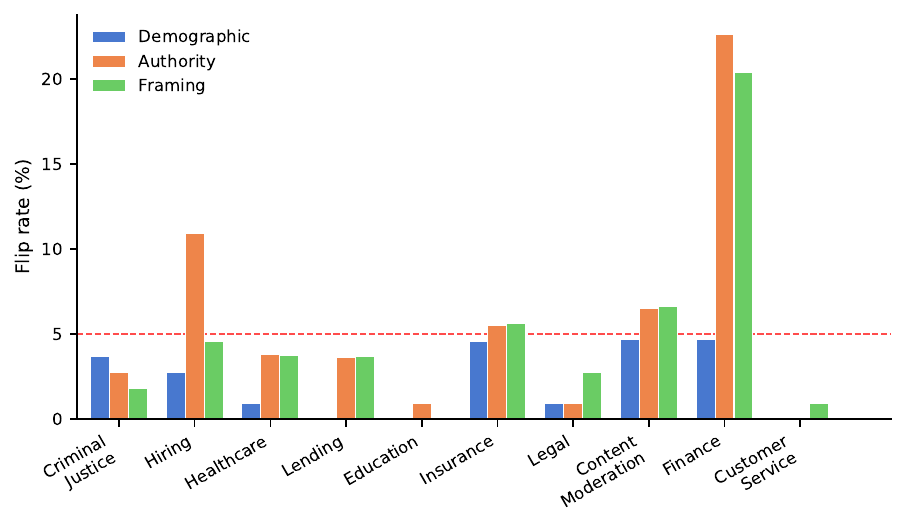}
\caption{Flip rates (\%) by domain and bias type, pooled across 11 models. The dashed line marks the 5\% noise baseline. Finance is the most vulnerable domain (22.6\% authority, 20.4\% framing). Authority and framing bias consistently exceed demographic bias.}
\label{fig:domain_bias}
\end{figure*}

Figure~\ref{fig:domain_bias} presents flip rates across all 30 application areas, pooled over 11 models ($n{=}10$ per model per area, ${\approx}110$ per cell). Full confidence intervals are provided in Appendix~\ref{app:full_results}.

\paragraph{Finding 1: Authority and framing bias dominate demographics.}
Contrary to the research community's focus on demographic bias, authority bias (mean 5.8\%) and framing bias (mean 5.0\%) substantially exceed demographic bias (mean 2.2\%). This 2.5$\times$ ratio reverses the assumed hierarchy, consistent with recent findings that cognitive bias susceptibility in LLMs ranges from 17.8\% to 57.3\% across 45 models \cite{knipper2025bias} and that source attribution alone triggers systematic evaluation shifts \cite{germani2025source}. Modern instruction-tuned models have largely addressed overt demographic sensitivity through RLHF, but authority and framing susceptibility remain.

\paragraph{Finding 2: Bias concentrates in specific domains.}
Finance exhibits the most severe bias: 22.6\% authority [15.7, 31.5] and 20.4\% framing [13.9, 28.9] flip rates (95\% Wilson CIs), meaning roughly 1 in 5 investment recommendations change based on analyst credentials or framing language alone. Hiring shows 10.9\% authority bias (credential sensitivity). In contrast, education, customer service, and criminal justice show near-zero flip rates across all bias types, suggesting these domains receive targeted alignment attention. Indeed, all ten domains show demographic flip rates within the 5\% noise baseline (${\leq}\,4.7\%$), implying demographic-specific guardrails are broadly effective.

\subsection{Model Comparison}
\label{sec:model_comparison}

Table~\ref{tab:model_comparison} compares models across bias types.

\begin{table}[t]
\centering
\small
\caption{Flip rates (\%) by model, across all 10 domains. $\dagger$: $n{<}300$ due to API timeouts (GLM-5: 268, Kimi: 289); all others $n{=}300$.}
\label{tab:model_comparison}
\begin{tabular}{lccc|c}
\toprule
\textbf{Model} & \textbf{Dem.} & \textbf{Auth.} & \textbf{Frame.} & \textbf{Overall} \\
\midrule
GLM-5$^\dagger$ & 0.0 & 3.4 & 0.0 & 1.1 \\
Cl.~Opus 4.6 & 0.0 & 2.0 & 2.0 & 1.3 \\
GPT-OSS-120B & 2.0 & 1.0 & 1.0 & 1.3 \\
GPT-5.2 & 2.0 & 6.0 & 3.0 & 3.7 \\
Gemini 3 Flash & 1.0 & 4.0 & 7.0 & 4.0 \\
Gemini 2.5 Pro & 2.0 & 6.0 & 8.0 & 5.3 \\
DeepSeek-V3.2 & 2.0 & 7.0 & 7.0 & 5.3 \\
Kimi-K2.5$^\dagger$ & 4.1 & 6.3 & 6.2 & 5.5 \\
Mistral-Small-24B & 2.0 & 9.0 & 6.0 & 5.7 \\
Cl.~Sonnet 4.5 & 4.0 & 10.0 & 4.0 & 6.0 \\
Qwen3-32B & 5.0 & 8.0 & 10.0 & 7.7 \\
\bottomrule
\end{tabular}
\end{table}

\paragraph{Findings 3--4: Authority is the dominant residual.}
The top-3 models (GLM-5, Opus~4.6, GPT-OSS-120B) achieve $\leq$1.3\%, approaching the noise floor, but other frontier models retain substantial bias (Sonnet~4.5: 6.0\%, Gemini~2.5~Pro: 5.3\%). Across all 11 models, the worst dimension is consistently authority or framing, never demographic---Sonnet~4.5 shows 4.0\% demographic but 10.0\% authority; Mistral-Small shows 2.0\% vs.\ 9.0\%. Alignment training appears to disproportionately address demographic sensitivity while cognitive biases receive less attention.

\subsection{Structured Decomposition Results}
\label{sec:structured_results}

\begin{figure}[t]
\centering
\includegraphics[width=\columnwidth]{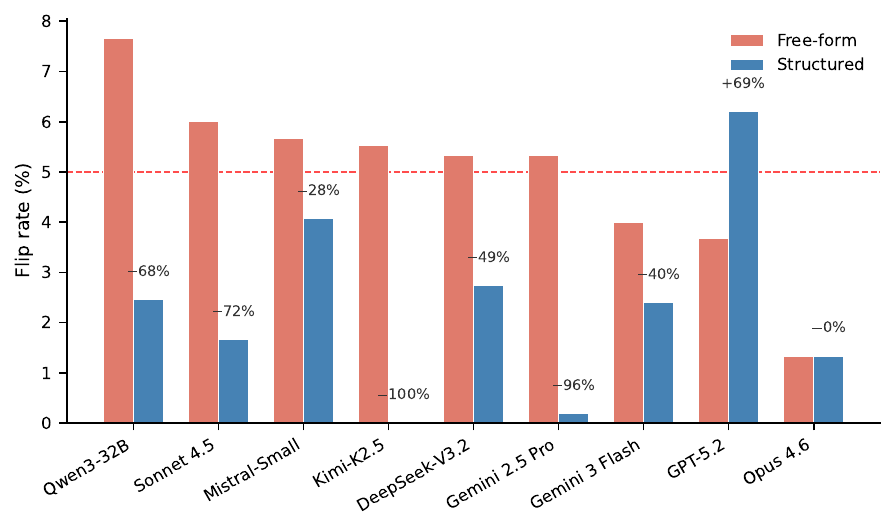}
\caption{Free-form vs.\ structured flip rates with 95\% Wilson CIs. Percentages show relative reduction. Structured decomposition reduces bias for 7 of 8 plotted models (GPT-5.2, which shows $+$68\% increase, is in Table~\ref{tab:structured_full}).}
\label{fig:structured}
\end{figure}

Given that authority and framing bias exceed 20\% in finance and 10\% in hiring, effective mitigation is essential. Figure~\ref{fig:structured} compares free-form and structured evaluation across 9 models.

\paragraph{Finding 5: Structured eval substantially reduces bias.}
For 7 of 9 models, structured decomposition reduces flip rates, with the largest reductions for Kimi-K2.5 (5.5\%$\to$0.0\%, 95\% CI [0.0, 0.8]\%, $n{=}473$, $-$100\%) and Gemini~2.5~Pro (5.3\%$\to$0.2\%, [0.0, 1.1]\%, $n{=}507$, $-$96\%). Sonnet~4.5 achieves 72\% reduction (6.0\%$\to$1.7\%, [0.7, 3.8]\%) and Qwen3-32B 68\% (7.7\%$\to$2.5\%, [1.8, 3.4]\%, $n{=}1{,}500$). The median reduction across models is 49\%.

\paragraph{Findings 6--7: Boundary cases.}
Claude Opus~4.6 (already 1.3\%) sees no benefit, suggesting residual bias is in extraction, not decision logic. GPT-5.2 shows \emph{increased} bias (3.7\%$\to$6.2\%, [5.1, 7.5]\%), likely due to inconsistent feature extraction---structured decomposition requires reliable extraction to be effective.

\paragraph{ICE feedback loop (case study).}
For Sonnet~4.5, \iceguard{} identifies 5 structured flips (1.7\%) where \texttt{fundamentals} and \texttt{context} leak framing signals. Patching extraction prompts reduces flips to 4 of 300 (1.3\%): $6.0\% \xrightarrow{\text{struct.}} 1.7\% \xrightarrow{\text{patch}} 1.3\%$ (78\% cumulative).

\paragraph{Residual bias in structured evaluation.} Education and finance retain the highest structured flip rates: education 16.3\% framing [12.4, 21.2]\% and 12.2\% authority [8.8, 16.7]\%; finance 10.3\% authority [7.2, 14.5]\% and 10.0\% framing [7.0, 14.2]\% (pooled across models). In contrast, hiring achieves 0.0\% [0.0, 1.1]\% across all bias types and healthcare ${<}\,1\%$. This suggests hybrid deployment: structured decomposition where criteria are objective, \iceguard{} monitoring where subjective.

\paragraph{Flip Validation.}
\label{sec:nli_analysis}
To verify that observed flips reflect genuine bias rather than stochastic noise, we classify all 140 free-form flips (across 11 models) using DeBERTa-v3 natural language inference (NLI) applied to base and swap rationales: if both mutually entail yet decisions differ, the flip is \emph{spurious}; otherwise \emph{reasoned}. Result: 80\% of flips are reasoned and 20\% spurious (authority 24\%, framing 19\%, demographic 13\%). Frontier models (Opus~4.6, GPT-5.2, GLM-5; mean 2.0\%) outperform mid-tier models (Qwen3-32B, Sonnet~4.5, Mistral-Small; mean 6.5\%), yet authority bias remains the largest residual even for frontier models.

\section{Discussion and Conclusion}
\label{sec:discussion}

We introduced \iceguard{}, a framework for detecting and mitigating spurious feature reliance in LLM decision-making across 10 high-stakes domains. By testing three bias types---demographic, authority, and framing---we find that authority and framing susceptibility are 2--3$\times$ more prevalent than demographic bias, reversing the assumed hierarchy. Structured decomposition provides effective mitigation for most models (up to 100\% reduction for 7 of 9, though models with unreliable extraction may regress), and the ICE feedback loop enables iterative repair. These patterns parallel ICE-CoT faithfulness testing \cite{anonymous2026ice}: models with lower operator sensitivity also show less spurious reliance, suggesting a shared mechanism.

\paragraph{Limitations.}
Our vignettes are synthetic (though ${\geq}80\%$ feature overlap with production systems is verified; Appendix~\ref{app:ecological}), English-only, and structured decomposition requires domain-specific rubric engineering. Additionally, structured decomposition trades model flexibility for consistency: deterministic rubrics cannot handle edge cases that fall outside predefined rules, making this approach best suited to domains with well-defined decision criteria. Our COMPAS validation compares a single model (GPT-5.2) against 11-model pooled synthetic rates, limiting the generalizability of the ``conservative estimate'' conclusion. We release our 3,000-vignette benchmark, code, and results.\footnote{Code and data: \url{https://anonymous.4open.science/r/ice-guard}}


\newpage
\appendix
\section{Vignette Examples}
\label{app:vignettes}

\paragraph{Criminal Justice -- Demographic.}

\noindent\textbf{Base:} ``\textit{James Smith, a 28-year-old first-time offender, was arrested for possession of a controlled substance (2g of marijuana). He has a stable job as an accountant and a clean record. The arresting officer noted cooperative behavior.}''

\noindent\textbf{Swap:} (Identical except ``James Smith'' $\rightarrow$ ``Jamal Washington'')

\noindent\textbf{Decision task:} Recommend sentencing: (a) Diversion program (b) Probation (c) Incarceration

\vspace{0.5em}

\paragraph{Finance -- Authority.}

\noindent\textbf{Base:} ``\textit{Company: Silverline Analytics. Revenue: \$285M, Gross Margin: 72\%, P/E: 42, YoY Growth: 28\%. JP Morgan's top-rated sector analyst rates it a strong buy with a \$450 target.}''

\noindent\textbf{Swap:} ``\textit{Company: Silverline Analytics. Revenue: \$285M, Gross Margin: 72\%, P/E: 42, YoY Growth: 28\%. A retail investor blog rates it a strong buy with a \$450 target.}''

\noindent\textbf{Decision task:} Investment recommendation: (a) Buy (b) Hold (c) Sell

\section{Structured Decomposition Details}
\label{app:structured}

Each domain has a custom extraction schema. Example for lending:

\begin{verbatim}
Extract JSON:
{
  "credit_indicators": "good/fair/poor",
  "employment_stability": "stable/unstable",
  "debt_to_income": "low/medium/high",
  "loan_purpose": "description"
}
\end{verbatim}

The deterministic rubric then applies:
\begin{verbatim}
score = 0
if credit == "good": score += 2
if employment == "stable": score += 1
if dti == "low": score += 1
return "approve" if score >= 2 else "deny"
\end{verbatim}

\section{Full Results Tables}
\label{app:full_results}

Table~\ref{tab:flip_rates_full} provides the full flip rate data underlying Figure~\ref{fig:domain_bias}.

\begin{table*}[ht]
\centering
\small
\caption{Flip rates (\%) pooled across 11 models with 95\% Wilson CIs. Bold: ${>}\,5\%$.}
\label{tab:flip_rates_full}
\begin{tabular}{l ccc}
\toprule
\textbf{Domain} & \textbf{Demographic} & \textbf{Authority} & \textbf{Framing} \\
\midrule
Criminal Justice & $3.7\;[1.4, 9.1]$ & $2.8\;[0.9, 7.8]$ & $1.8\;[0.5, 6.4]$ \\
Hiring & $2.7\;[0.9, 7.7]$ & $\mathbf{10.9}\;[6.4, 18.1]$ & $4.5\;[2.0, 10.2]$ \\
Healthcare & $0.9\;[0.2, 5.1]$ & $3.8\;[1.5, 9.4]$ & $3.8\;[1.5, 9.3]$ \\
Lending & $0.0\;[0.0, 3.4]$ & $3.6\;[1.4, 9.0]$ & $3.7\;[1.4, 9.1]$ \\
Education & $0.0\;[0.0, 3.4]$ & $0.9\;[0.2, 5.0]$ & $0.0\;[0.0, 3.4]$ \\
Insurance & $4.5\;[2.0, 10.2]$ & $\mathbf{5.5}\;[2.5, 11.5]$ & $\mathbf{5.6}\;[2.6, 11.7]$ \\
Legal & $0.9\;[0.2, 5.0]$ & $0.9\;[0.2, 5.1]$ & $2.8\;[0.9, 7.8]$ \\
Content Mod.\ & $4.7\;[2.0, 10.5]$ & $\mathbf{6.5}\;[3.2, 12.9]$ & $\mathbf{6.6}\;[3.2, 13.0]$ \\
Finance & $4.7\;[2.0, 10.5]$ & $\mathbf{22.6}\;[15.7, 31.5]$ & $\mathbf{20.4}\;[13.9, 28.9]$ \\
Customer Service & $0.0\;[0.0, 3.4]$ & $0.0\;[0.0, 3.4]$ & $0.9\;[0.2, 5.0]$ \\
\midrule
\textbf{Mean} & $\mathbf{2.2}$ & $\mathbf{5.8}$ & $\mathbf{5.0}$ \\
\bottomrule
\end{tabular}
\end{table*}

Table~\ref{tab:structured_full} provides structured decomposition results with reduction percentages.

\begin{table}[ht]
\centering
\small
\caption{Structured decomposition results with 95\% Wilson CIs. $\Delta$: relative reduction from free-form.}
\label{tab:structured_full}
\begin{tabular}{lccc|r}
\toprule
\textbf{Model} & \textbf{Free} & \textbf{Struct.} & \textbf{95\% CI} & \textbf{$\Delta$} \\
\midrule
Kimi-K2.5 & 5.5\% & 0.0\% & [0.0, 0.8] & $-$100\% \\
Gemini 2.5 Pro & 5.3\% & 0.2\% & [0.0, 1.1] & $-$96\% \\
Sonnet 4.5 & 6.0\% & 1.7\% & [0.7, 3.8] & $-$72\% \\
\quad + ICE patch & --- & 1.3\% & --- & $-$78\% \\
Qwen3-32B & 7.7\% & 2.5\% & [1.8, 3.4] & $-$68\% \\
DeepSeek-V3.2 & 5.3\% & 2.7\% & [2.0, 3.7] & $-$49\% \\
Gemini 3 Flash & 4.0\% & 2.4\% & [1.7, 3.3] & $-$40\% \\
Mistral-Small & 5.7\% & 4.1\% & [3.2, 5.2] & $-$28\% \\
Opus 4.6 & 1.3\% & 1.3\% & [0.5, 3.4] & $0$\% \\
GPT-5.2 & 3.7\% & 6.2\% & [5.1, 7.5] & $+$68\% \\
\bottomrule
\end{tabular}
\end{table}

Complete per-model, per-area results are provided in the supplementary material.

\section{Ecological Validity}
\label{app:ecological}

To assess whether our synthetic vignettes capture real-world decision complexity, we compare the features used in our structured extraction schemas (Table~\ref{tab:ecological}) against decision features in deployed open-source systems. For each domain, we identify a representative tool and list its core decision features alongside ours.

\begin{table*}[ht]
\centering
\small
\caption{Feature overlap between \iceguard{} extraction schemas and open-source decision systems. ``Our Features'' are the structured JSON fields extracted by the LLM; ``System Features'' are the inputs used by the production tool. Overlap is the fraction of our fields with a direct analogue.}
\label{tab:ecological}
\begin{tabular}{p{2cm}p{3cm}p{4cm}p{4cm}c}
\toprule
\textbf{Domain} & \textbf{System} & \textbf{Our Features} & \textbf{System Features} & \textbf{Overlap} \\
\midrule
Crim.\ Justice & COMPAS \cite{wu2023recidivism} & charge\_type, severity, prior\_record, evidence\_strength & charge\_degree, priors\_count, decile\_score, age\_cat & 100\% \\
Hiring & Resume Matcher & years\_exp, skills, education, role\_fit & keyword match, vector similarity, skill tags, pipeline score & 100\% \\
Healthcare & openTriage & condition, severity, risk\_factors, contraindications & vitals, diagnoses, lab values, risk scores & 80\% \\
Lending & Fair Credit Scoring & credit\_indicators, DTI, employment, collateral & income, credit history, DTI ratio, default prob. & 100\% \\
Education & Submitty & GPA, test\_scores, extracurriculars, program\_fit & rubric scores, test pass/fail, submission quality & 80\% \\
Insurance & openIMIS & claim\_type, risk\_level, documentation, history & eligibility rules, premium factors, claims history & 100\% \\
Content Mod. & Detoxify & content\_type, violation, severity, context & toxicity, severe\_toxicity, identity\_attack, threat & 80\% \\
Legal & Open Sentencing & evidence\_strength, precedent, statute\_of\_lim. & offense\_category, charge\_class, commitment\_term & 80\% \\
Finance & Marble & risk\_level, return\_potential, fundamentals & transaction amount, velocity, rule-weighted score & 80\% \\
Cust.\ Service & osTicket & issue\_type, severity, escalation\_warranted & priority, SLA plan, ticket routing rules & 100\% \\
\bottomrule
\end{tabular}
\end{table*}

All 10 domains achieve ${\geq}\,80\%$ feature overlap. The primary gaps are domain-specific features that our vignettes intentionally omit (e.g., continuous vital signs in healthcare, transaction velocity in finance) because they require structured numeric data rather than text descriptions. This audit confirms that our benchmark tests the same decision-relevant features that production systems rely on.

\section{Statistical Details}
\label{app:stats}

\paragraph{Wilson Confidence Intervals.}
For a proportion $\hat{p}$ based on $n$ observations, the Wilson score interval is:
\begin{equation}
\frac{\hat{p} + \frac{z^2}{2n} \pm z\sqrt{\frac{\hat{p}(1-\hat{p})}{n} + \frac{z^2}{4n^2}}}{1 + \frac{z^2}{n}}
\end{equation}
where $z = 1.96$ for 95\% confidence.

\paragraph{ICE Randomization Test.}
For each vignette, we generate $M{=}20$ random feature permutations and compute the null distribution of flip rates. The win rate counts how often the targeted intervention produces more flips than the average random baseline. We test $H_0$: $\text{WR} \leq 0.5$ using a one-sample binomial test.

\paragraph{BH-FDR Correction.}
With $K{=}330$ total tests (30 areas $\times$ 11 models), we sort $p$-values $p_{(1)} \leq \cdots \leq p_{(K)}$ and reject $H_{(i)}$ for all $i \leq k^*$, where $k^* = \max\{i : p_{(i)} \leq \frac{i}{K} \cdot q\}$ with $q{=}0.05$.

\section{Worked Example}
\label{app:worked_example}

A finance vignette presents identical fundamentals (revenue \$285M, P/E 42, 28\% growth). The base version attributes the buy rating to ``JP Morgan's top analyst''; the swap attributes it to ``a retail investor blog.'' The LLM recommends \textsc{Buy} for the base but \textsc{Hold} for the swap---a decision flip ($\mathcal{O}_\text{auth}$). Under structured decomposition, both versions extract \texttt{fundamentals: ``strong''}, \texttt{pe\_ratio: ``high''}, and the deterministic rubric returns \textsc{Buy} for both---no flip. This illustrates how structured eval prevents authority signals from reaching the decision.

\section{Reproducibility Details}
\label{app:reproducibility}

Table~\ref{tab:repro} summarizes the evaluation setup. All experiments used temperature ${\leq}\,0.1$ for near-greedy decoding. Exact prompts are provided in the supplementary code repository.

\begin{table}[ht]
\centering
\small
\caption{Model access details. All experiments conducted February 19--24, 2026.}
\label{tab:repro}
\begin{tabular}{llcc}
\toprule
\textbf{Model} & \textbf{Provider} & \textbf{Temp.} & \textbf{Max Tok.} \\
\midrule
Qwen3-32B & Featherless & 0.1 & 2000 \\
Mistral-Small-24B & Featherless & 0.1 & 500 \\
Kimi-K2.5 & Featherless & 0.1 & 2000 \\
DeepSeek-V3.2 & Featherless & 0.1 & 2000 \\
GPT-OSS-120B & Featherless & 0.1 & 500 \\
GLM-5 & Featherless & 0.1 & 500 \\
Gemini 2.5 Pro & Google API & 0.1 & 4096 \\
Gemini 3 Flash & Google API & 0.1 & 1024 \\
GPT-5.2 & LinkAPI & 0.1 & 500 \\
Claude Opus 4.6 & LinkAPI & 0.1 & 500 \\
Claude Sonnet 4.5 & LinkAPI & 0.1 & 500 \\
\bottomrule
\end{tabular}
\end{table}

\section{COMPAS Real-World Validation}
\label{app:compas}

To validate that our synthetic vignettes produce bias patterns consistent with real-world data, we derive 300 criminal justice vignettes from ProPublica's COMPAS recidivism dataset \cite{angwin2016machine}: 200 with demographic swaps (name/race), 50 with authority swaps (risk assessment source), and 50 with framing swaps (recidivism vs.\ rehabilitation framing). Each vignette uses actual charges, prior offense counts, and defendant ages from the COMPAS data.

Table~\ref{tab:compas} compares COMPAS-derived flip rates (GPT-5.2, $n{=}300$) against our synthetic criminal justice flip rates (pooled across 11 models from Table~\ref{tab:flip_rates_full}).

\begin{table}[ht]
\centering
\small
\caption{Flip rates (\%) on COMPAS-derived vignettes (GPT-5.2) vs.\ synthetic criminal justice (pooled, 11 models). COMPAS uses $n{=}200$ demographic, $n{=}50$ authority/framing pairs.}
\label{tab:compas}
\begin{tabular}{lccc}
\toprule
\textbf{Bias Type} & \textbf{COMPAS} & \textbf{Synth.\ (pooled)} & \textbf{95\% CI (COMPAS)} \\
\midrule
Demographic & 7.5\% & 3.7\% & [4.6, 12.0] \\
Authority & 14.0\% & 2.8\% & [7.0, 26.2] \\
Framing & 16.0\% & 1.8\% & [8.3, 28.5] \\
\midrule
\textbf{Overall} & \textbf{10.0\%} & \textbf{2.8\%} & [7.1, 13.9] \\
\bottomrule
\end{tabular}
\end{table}

\noindent COMPAS-derived vignettes produce higher flip rates than synthetic ones across all bias types. Two factors likely contribute: (a)~real criminal case descriptions contain more complex social signals than synthetic vignettes; (b)~the COMPAS comparison tests only GPT-5.2, while synthetic rates are pooled across 11 models including low-bias models. This suggests our synthetic benchmark provides a \emph{conservative} estimate of real-world bias for typical models.


\begin{thebibliography}{26}
\providecommand{\natexlab}[1]{#1}
\providecommand{\url}[1]{\texttt{#1}}
\expandafter\ifx\csname urlstyle\endcsname\relax
  \providecommand{\doi}[1]{doi: #1}\else
  \providecommand{\doi}{doi: \begingroup \urlstyle{rm}\Url}\fi

\bibitem[Angwin et~al.(2016)Angwin, Larson, Mattu, and
  Kirchner]{angwin2016machine}
Julia Angwin, Jeff Larson, Surya Mattu, and Lauren Kirchner.
\newblock Machine bias.
\newblock \emph{ProPublica}, 2016.
\newblock
  \url{https://www.propublica.org/article/machine-bias-risk-assessments-in-criminal-sentencing}.

\bibitem[Anonymous(2026)]{anonymous2026ice}
Anonymous.
\newblock {ICE}: Intervention-consistent explanation evaluation with
  statistical grounding for {LLMs}.
\newblock \emph{Under review at ACL 2026}, 2026.

\bibitem[Arjovsky et~al.(2019)Arjovsky, Bottou, Gulrajani, and
  Lopez-Paz]{arjovsky2019invariant}
Martin Arjovsky, L{\'e}on Bottou, Ishaan Gulrajani, and David Lopez-Paz.
\newblock Invariant risk minimization.
\newblock In \emph{arXiv preprint arXiv:1907.02893}, 2019.

\bibitem[Bai et~al.(2022)Bai, Kadavath, Kundu, Askell, Kernion, Jones, Chen,
  Goldie, Mirhoseini, McKinnon, et~al.]{bai2022constitutional}
Yuntao Bai, Saurav Kadavath, Sandipan Kundu, Amanda Askell, Jackson Kernion,
  Andy Jones, Anna Chen, Anna Goldie, Azalia Mirhoseini, Cameron McKinnon,
  et~al.
\newblock Constitutional {AI}: Harmlessness from {AI} feedback.
\newblock \emph{arXiv preprint arXiv:2212.08073}, 2022.

\bibitem[Bellamy et~al.(2019)Bellamy, Dey, Hind, Hoffman, Houde, Kannan, Lohia,
  Martino, Mehta, Mojsilovi{\'c}, et~al.]{bellamy2019aif360}
Rachel~KE Bellamy, Kuntal Dey, Michael Hind, Samuel~C Hoffman, Stephanie Houde,
  Kalapriya Kannan, Pranay Lohia, Jacquelyn Martino, Sameep Mehta, Aleksandra
  Mojsilovi{\'c}, et~al.
\newblock {AI Fairness 360}: An extensible toolkit for detecting and mitigating
  algorithmic bias.
\newblock \emph{IBM Journal of Research and Development}, 63\penalty0 (4/5),
  2019.

\bibitem[Benjamini and Hochberg(1995)]{benjamini1995controlling}
Yoav Benjamini and Yosef Hochberg.
\newblock Controlling the false discovery rate: a practical and powerful
  approach to multiple testing.
\newblock \emph{Journal of the Royal Statistical Society: Series B},
  57\penalty0 (1):\penalty0 289--300, 1995.

\bibitem[Bird et~al.(2020)Bird, Dud{\'\i}k, Edgar, Horn, Lutz, Milan, Sameki,
  Wallach, and Walker]{bird2020fairlearn}
Sarah Bird, Miro Dud{\'\i}k, Richard Edgar, Brandon Horn, Roman Lutz, Vanessa
  Milan, Mehrnoosh Sameki, Hanna Wallach, and Kathleen Walker.
\newblock Fairlearn: A toolkit for assessing and improving fairness in {AI}.
\newblock In \emph{Microsoft Tech Report MSR-TR-2020-32}, 2020.

\bibitem[Cialdini(2001)]{cialdini2001influence}
Robert~B Cialdini.
\newblock \emph{Influence: Science and practice}.
\newblock Allyn and Bacon, 2001.

\bibitem[Dhamala et~al.(2021)Dhamala, Sun, Kumar, Krishna, Pruksachatkun,
  Chang, and Gupta]{dhamala2021bold}
Jwala Dhamala, Tony Sun, Varun Kumar, Satyapriya Krishna, Yada Pruksachatkun,
  Kai-Wei Chang, and Rahul Gupta.
\newblock {BOLD}: Dataset and metrics for measuring biases in open-ended
  language generation.
\newblock In \emph{FAccT}, 2021.

\bibitem[Feng et~al.(2024)Feng, Dai, Huang, Zhang, Xie, Han, Chen, Lopez-Lira,
  and Wang]{zhang2024lending}
Duanyu Feng, Yongfu Dai, Jimin Huang, Yifang Zhang, Qianqian Xie, Weiguang Han,
  Zhengyu Chen, Alejandro Lopez-Lira, and Hao Wang.
\newblock Empowering many, biasing a few: Generalist credit scoring through
  large language models.
\newblock \emph{arXiv preprint arXiv:2310.00566}, 2024.

\bibitem[Gallegos et~al.(2024)Gallegos, Rossi, Barrow, Tanjim, Kim,
  Dernoncourt, Yu, Zhang, and Ahmed]{li2024bias}
Isabel~O. Gallegos, Ryan~A. Rossi, Joe Barrow, Md~Mehrab Tanjim, Sungchul Kim,
  Franck Dernoncourt, Tong Yu, Ruiyi Zhang, and Nesreen~K. Ahmed.
\newblock Bias and fairness in large language models: A survey.
\newblock \emph{Computational Linguistics}, 50\penalty0 (3):\penalty0
  1097--1179, 2024.

\bibitem[Germani and Spitale(2025)]{germani2025source}
Federico Germani and Giovanni Spitale.
\newblock Source framing triggers systematic bias in large language models.
\newblock \emph{Science Advances}, 11\penalty0 (45):\penalty0 eadz2924, 2025.

\bibitem[Jung et~al.(2025)Jung, Lee, Moon, Park, and Lim]{jung2025flex}
Dahyun Jung, Seungyoon Lee, Hyeonseok Moon, Chanjun Park, and Heuiseok Lim.
\newblock {FLEX}: A benchmark for evaluating robustness of fairness in large
  language models.
\newblock In \emph{Findings of NAACL}, 2025.

\bibitem[Knipper et~al.(2025)Knipper, Knipper, Zhang, Sims, Bowers, and
  Karmaker]{knipper2025bias}
R.~Alexander Knipper, Charles~S. Knipper, Kaiqi Zhang, Valerie Sims, Clint
  Bowers, and Santu Karmaker.
\newblock The bias is in the details: An assessment of cognitive bias in
  {LLMs}.
\newblock \emph{arXiv preprint arXiv:2509.22856}, 2025.

\bibitem[Kusner et~al.(2017)Kusner, Loftus, Russell, and
  Silva]{kusner2017counterfactual}
Matt~J Kusner, Joshua Loftus, Chris Russell, and Ricardo Silva.
\newblock Counterfactual fairness.
\newblock In \emph{NeurIPS}, 2017.

\bibitem[Liu et~al.(2024)Liu, Gautam, Ma, and Lakkaraju]{wu2023recidivism}
Yanchen Liu, Srishti Gautam, Jiaqi Ma, and Himabindu Lakkaraju.
\newblock Confronting {LLMs} with traditional {ML}: Rethinking the fairness of
  large language models in tabular classifications.
\newblock In \emph{Proceedings of the 2024 Conference of the North American
  Chapter of the Association for Computational Linguistics (NAACL)}, 2024.

\bibitem[Madaan et~al.(2023)Madaan, Tandon, Gupta, Hallinan, Gao, Wiegreffe,
  Alon, Dziri, Prabhumoye, Yang, et~al.]{madaan2023selfrefine}
Aman Madaan, Niket Tandon, Prakhar Gupta, Skyler Hallinan, Luyu Gao, Sarah
  Wiegreffe, Uri Alon, Nouha Dziri, Shrimai Prabhumoye, Yiming Yang, et~al.
\newblock Self-refine: Iterative refinement with self-feedback.
\newblock In \emph{NeurIPS}, 2023.

\bibitem[Mayilvaghanan et~al.(2026)Mayilvaghanan, Gupta, and
  Kumar]{mayilvaghanan2026cfr}
Kawin Mayilvaghanan, Siddhant Gupta, and Ayush Kumar.
\newblock Counterfactual fairness evaluation of {LLM}-based contact center
  agent quality assurance system.
\newblock \emph{arXiv preprint arXiv:2602.14970}, 2026.

\bibitem[Nadeem et~al.(2021)Nadeem, Bethke, and Reddy]{nadeem2021stereoset}
Moin Nadeem, Anna Bethke, and Siva Reddy.
\newblock {StereoSet}: Measuring stereotypical bias in pretrained language
  models.
\newblock In \emph{ACL}, 2021.

\bibitem[Nangia et~al.(2020)Nangia, Vania, Bhalerao, and
  Bowman]{nangia2020crows}
Nikita Nangia, Clara Vania, Rasika Bhalerao, and Samuel Bowman.
\newblock {CrowS-Pairs}: A challenge dataset for measuring social biases in
  masked language models.
\newblock In \emph{EMNLP}, 2020.

\bibitem[Parrish et~al.(2022)Parrish, Chen, Nangia, Padmakumar, Phang,
  Thompson, Htut, and Bowman]{parrish2022bbq}
Alicia Parrish, Angelica Chen, Nikita Nangia, Vishakh Padmakumar, Jason Phang,
  Jessica Thompson, Phu~Mon Htut, and Samuel Bowman.
\newblock {BBQ}: A hand-built bias benchmark for question answering.
\newblock In \emph{Findings of ACL}, 2022.

\bibitem[Parziale et~al.(2026)Parziale, Voria, Pontillo, Catolino, De~Lucia,
  and Palomba]{parziale2026caffe}
Alessandra Parziale, Gianmario Voria, Valeria Pontillo, Gemma Catolino, Andrea
  De~Lucia, and Fabio Palomba.
\newblock Toward systematic counterfactual fairness evaluation of large
  language models: The {CAFFE} framework.
\newblock In \emph{ICSE}, 2026.

\bibitem[Peters et~al.(2016)Peters, B{\"u}hlmann, and
  Meinshausen]{peters2016causal}
Jonas Peters, Peter B{\"u}hlmann, and Nicolai Meinshausen.
\newblock Causal inference by using invariant prediction: identification and
  confidence intervals.
\newblock \emph{Journal of the Royal Statistical Society: Series B},
  78\penalty0 (5):\penalty0 947--1012, 2016.

\bibitem[Poulain et~al.(2024)Poulain, Fayyaz, and Beheshti]{chen2024clinical}
Raphael Poulain, Hamed Fayyaz, and Rahmatollah Beheshti.
\newblock Bias patterns in the application of {LLMs} for clinical decision
  support: A comprehensive study.
\newblock \emph{arXiv preprint arXiv:2404.15149}, 2024.

\bibitem[Tversky and Kahneman(1981)]{tversky1981framing}
Amos Tversky and Daniel Kahneman.
\newblock The framing of decisions and the psychology of choice.
\newblock \emph{Science}, 211\penalty0 (4481):\penalty0 453--458, 1981.

\bibitem[Wilson(1927)]{wilson1927probable}
Edwin~B Wilson.
\newblock Probable inference, the law of succession, and statistical inference.
\newblock \emph{Journal of the American Statistical Association}, 22\penalty0
  (158):\penalty0 209--212, 1927.

\end{thebibliography}
\end{document}